\pdfoutput=1

\documentclass[11pt]{article}

\usepackage{EACL2023}

\usepackage{times}
\usepackage{latexsym}

\usepackage[normalem]{ulem}
\usepackage{enumitem}
\usepackage{graphicx}
\usepackage{amsmath}
\usepackage{amssymb}
\usepackage{algorithm}
\usepackage[noend]{algpseudocode}
\usepackage{caption}
\usepackage{subcaption}
\usepackage{float}
\usepackage{colortbl}

\usepackage[T1]{fontenc}

\usepackage[utf8]{inputenc}

\usepackage{microtype}

\usepackage{inconsolata}

\title{Pento-D\textsc{ia}Ref: A Diagnostic Dataset for Learning the Incremental\\Algorithm for Referring Expression Generation from Examples}

\author{Philipp Sadler$^\mathbf{1}$ {\normalfont and} David Schlangen$^\mathbf{1,2}$ \\
  $^\mathbf{1}$CoLabPotsdam / Computational Linguistics \\
  Department of Linguistics, University of Potsdam, Germany \\
  $^\mathbf{2}$German Research Center for Artificial Intelligence (DFKI), Berlin, Germany \\
  \texttt{firstname.lastname@uni-potsdam.de}\\}

\newcommand{\datasetname}{Pento-D\textsc{ia}Ref}

\begin{document}
\maketitle
\begin{abstract}

NLP tasks are typically defined extensionally through datasets containing example instantiations (e.g., pairs of image $i$ and text $t$), but motivated intensionally through capabilities invoked in verbal descriptions of the task (e.g., ``$t$ is a description of $i$, for which the content of $i$ needs to be recognised and understood'').
We present \datasetname, a diagnostic dataset in a visual domain of puzzle pieces where referring expressions are generated by a well-known symbolic algorithm (the ``Incremental Algorithm''),
which itself is motivated by appeal to a hypothesised capability (eliminating distractors through application of Gricean maxims). Our question then is whether the extensional description (the dataset) is sufficient for a neural model to pick up the underlying regularity and exhibit this capability given the simple task definition of producing expressions from visual inputs. We find that a model supported by a vision detection step and a targeted data generation scheme 
achieves an almost perfect BLEU@1 score and sentence accuracy, whereas simpler baselines do not.
\end{abstract}

\section{Introduction}

\begin{figure}[t]
    \begin{center}
        \includegraphics[width=0.43\textwidth]{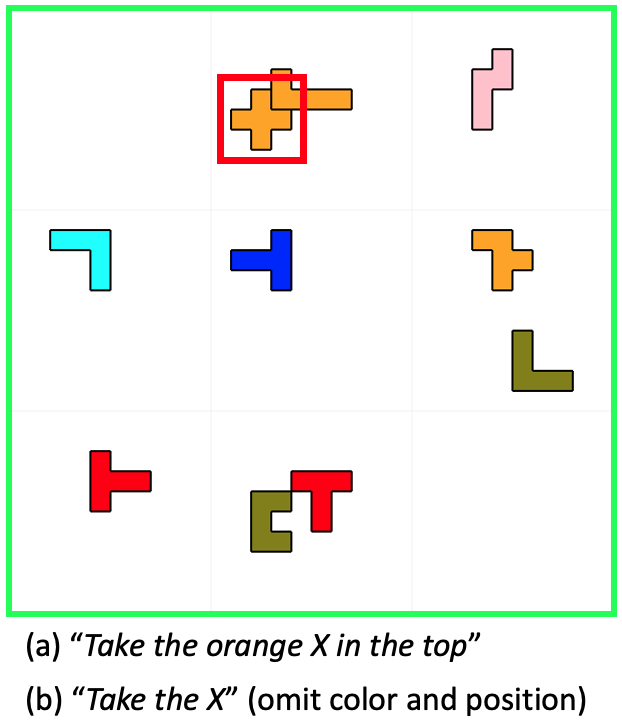}
    \end{center}
    \vspace{-0.1cm}
    \caption{An example board with a referring
expression (b) as produced by the Incremental Algorithm (\textsc{ia}) (minimal wrt. a preference order); and an unnecessarily verbose reference (a) (that still is uniquely referring). The reference target is highlighted with bounding box; images regions separated by addition of lines.}
    \label{fig:example_board}
   \vspace{-0.5cm}
\end{figure}

Being able to effectively and efficiently \textit{refer} to objects is a central component of human language competence
\citep{deemter_2016}. The computational task of referring expression generation (REG) goes beyond the production of %
image descriptions (as in image captioning), in that it is a \textit{uniquely identifying} description that needs to be produced, given a specific situation.
In the formulation of \citet{krahmer-van-deemter-2012-computational}, the REG task involves reasoning over all relevant objects in a scene, in order to determine what would make a description uniquely identifying.
Additionally, maxims of efficiency \citep{grice_logic_1967} predict that it is a \textit{minimal} natural language expression that should be preferred. The \textit{Incremental Algorithm} (\textsc{ia}) \citep{dale_computational_1995} is a well-known classic symbolic algorithm that tries to realise these desiderata. For  example given a reference target and various distractors as in Figure~\ref{fig:example_board} (an example of the domain chosen in this paper (Pentomino, \citet{golomb_1996,zarriess-etal-2016-pentoref,Kennington-2017})), then the Incremental Algorithm (\textsc{ia}) produces ``Take the X'', achieving the desired uniquely identifying reference by mentioning only the shape and \textit{not} also color and position.
 
\begin{table*}[t]
\centering
\begin{small}
\begin{tabular}{|l|c|c|l|l|l|}
\hline
\textbf{Diagnostic Dataset}           & \textbf{Task} & \textbf{Input} & \textbf{Condition} & \textbf{Output} & \textbf{Generalizability Testing} \\ \hline
\citet{wu_reascan_2021}    & Nav.          & Symb. State   &  Text (Command)       & Text (Actions)  & Words, Phrases, Action Length        \\ \hline
\citet{liu_clevr-ref_2019} & REC           & Image          & Text (RE)              & BBox            & Color-Shapes                         \\ \hline
\datasetname\ (Ours)        & REG           & Image          & BBox               & Text (\textsc{ia}-RE)       & Color-Shapes, Positions, \textsc{ia}-REs \\ \hline
\end{tabular}
\end{small}
\caption{The most relevant datasets in comparison to \datasetname. In contrast to \citet{liu_clevr-ref_2019} we study the task of REG (which avoids models to exploit language inputs) and add generalization tests for the output expressions. 
}
\label{table:related_datasets}
\vspace{-0.4cm}
\end{table*}

Can such a reference strategy be learned by neural generation models from visual inputs alone? This is a question that is typically not systematically challenged in language generation from images \citep{referitgame, refcoco, refcocog,flickr, refclef}, as in natural scenes (such as in the RefCOCO dataset \citep{refcoco}), it has been shown that descriptions can be produced based on the recognition of only parts of the image \citep{agrawal_analyzing_2016}; our dataset is designed to make this impossible.
\citet{schlangen-2021-targeting} observed that in typical settings in the field of natural language processing, the connection between an underlying natural language capability and a learned model is only an indirect one. It rests on how well the dataset from which the model was induced does indeed exemplify the assumed underlying task---of which typically only a verbal description is given---and in turn on the extent to which the task represents the capability.

In this work we study how a intensionally defined task (in the distinction of \citet{schlangen-2021-targeting}) for which a verbal and theoretically motivated description is given (through a symbolic algorithm) can be learned from its extensional exemplification.
Our contention is that the use of synthetic data \citep{johnson_clevr_2017, liu_clevr-ref_2019, scan,gscan,wu_reascan_2021} offers the opportunity to strengthen the link, insofar as guarantees can be given on the exemplification relation. More specifically, we choose the  Incremental Algorithm \citep{dale_computational_1995} for the data generation process, which itself comes with a motivation through recourse to underlying fundamental conversational capabilities (appeal to Gricean maxims, \citet{grice_logic_1967}). Our contributions are as follows:\footnote{%
The source code and datasets are made publicly available at \url{https://github.com/clp-research/pento-diaref}.}
\begin{itemize}
  \itemsep-0.1cm
  \item We create a novel synthetic dataset, \datasetname, of examples that pairs visual scenes with generated referring expressions;
  \item examine two variants of the dataset, representing two different ways to exemplify the underlying task;
  \item and evaluate an LSTM-based baseline \citep{refcocog}, a transformer \citep{vaswani_attention_2017} and a modified version with region embeddings \citep{tan_lxmert_2019} on them.
\end{itemize}

\section{Related Work}
\label{sec:related_work}
\vspace{-1mm}
\paragraph{Compositional Reasoning.} \citet{scan} introduced a systematic benchmark to test the generalization capabilities of recurrent neural networks through the use of compositional splits and found that these models fail ``spectacularly''. \citet{gscan} extended the task of mapping text commands to actions (Navigation) by conditioning the learner additionally on a symbolic world state. Later \cite{wu_reascan_2021} 
provided a curated dataset along with new dimensions for generalizability testing. Our work follows the idea of  generalization testing through compositional datasets in language and vision settings where training examples are composed in such a way that the models are exposed towards all property values of objects, but not to all the possible combinations of them, so that they can be tested on unseen combinations. In contrast to their work we use images instead of symbolic world states as the input.

\vspace{-1mm}
\paragraph{Diagnostic Datasets.} For the generation of the synthetic data we took inspiration from \citet{johnson_clevr_2017}  who created a ``diagnostic dataset'' for visual question answering to test for model limitations. They draw 3D objects on a 2D plane and systematically use templates to create questions about the objects to avoid biases that occur in ``common'' datasets.
Later \citet{liu_clevr-ref_2019} convert the questions  to referring expressions to test systematically for referring expression comprehension (REC). They claim that the models' performance on the dataset proves that they ``work as intended''. In this work we study this aspect as well but on the mirroring task of REG which avoids models to exploit hints from the language inputs (Table~\ref{table:related_datasets}).
\vspace{-1mm}
\paragraph{Program Learners} As a related idea \citet{pi_reasoning_2022} suggest to train language models on text outputs of ``executable programs'' (which could be a symbolic algorithm). They focus on the pre-training paradigm and aim to induce reasoning capabilities into language models to enhance their usefulness for downstream tasks. Our work is more specifically focused on the question whether a neural model is able to learn the underlying capabilities that are exhibited by a symbolic algorithm in a vision and language domain. \citet{same_non-neural_2022} showed that such rule-based algorithms are still a useful approach for REG in natural settings.

\section{\datasetname\ Task and Dataset}
\label{sec:synthetic_reg}

We present a \textbf{D}iagnostic dataset of \textbf{\textsc{ia}} \textbf{Ref}erences in a \textbf{Pento}mino domain (\datasetname) that ties extensional and intensional definitions more closely together, insofar as the latter is the generative process creating the former \citep{schlangen-2021-targeting}. In this chapter we describe the task (§\ref{sec:task}) and how it is tied to the Incremental Algorithm (§\ref{sec:ia}) via the generation process (§\ref{sec:data_generation}) and present our compositional splits (§\ref{section:holdouts}) for generalization testing.

\subsection{Task Description} %
\label{sec:task}

\begin{figure}[t]
    \begin{center}
        \includegraphics[width=0.45\textwidth]{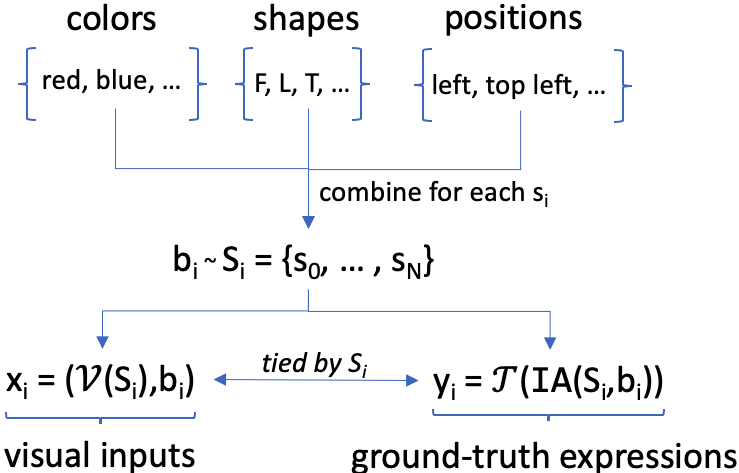}
    \end{center}
    \caption{The general generation process for our synthetic datasets as defined by the task.}
    \label{fig:task}
    \vspace{-0.3cm}
\end{figure}

Given as input an $x_i = (v_i, b_i)$, representing a Pentomino board $v_i$ as in Figure~\ref{fig:example_board} and a bounding box $b_i$ (indicating the target piece), a model $f$ has to produce a referring expression $y_i$ (as it would be generated by \textsc{IA}) as shown in Figure~\ref{fig:task}. Formally, this can be described either as a classification task $\text{argmax}_{y_i}P(y_i|x_i)$ when $y_i$ is considered a whole sentence or more generally as a conditional language modeling task $P(w_t|w_{<t},x_i)$ with $y_i = \{w_0,...,w_T\}$ where $T$ is the length of the expression. We present models for both of these interpretations in Section \ref{sec:models}.

\begin{algorithm}[b]
\caption{The \textsc{ia} on symbolic properties as based on the formulation by \citet{deemter_2016}}\label{alg:ia}
\begin{algorithmic}[1]
\Require{A set of distractors $M$, a set of property values $\mathcal{P}$ of a referent $r$ and a linear preference order $\mathcal{O}$ over the property values $\mathcal{P}$}
\State{$\mathcal{D} \gets \emptyset $}

\For{$P$ in $\mathcal{O}(\mathcal{P})$}
\State{$\mathcal{E} \gets \{ m \in M: \neg P(m)$\}}
\If{$\mathcal{E} \ne \emptyset$}
    \State Add $P$ to $\mathcal{D}$
    \State Remove $\mathcal{E}$ from $M$
\EndIf
\EndFor
\State{\textbf{return} $\mathcal{D}$}
\end{algorithmic}
\end{algorithm}

\subsection{The Incremental Algorithm (\textsc{ia})}
\label{sec:ia}

The Algorithm~\ref{alg:ia} , in the formulation of \cite{krahmer-van-deemter-2012-computational},
is  supposed to find the properties that uniquely identify an object among others given a preference over properties. To accomplish this the algorithm is given the property values $\mathcal{P}$ of distractors in $M$ and of a referent $r$. Then the algorithm excludes distractors in several iterations until either $M$ is empty or every property of $r$ has been tested. During the exclusion process the algorithm computes the set of distractors that do \textit{not} share a given property with the referent and stores the property in $\mathcal{D}$. These properties in $\mathcal{D}$ are the ones that distinguish the referent from the others and thus will be returned.

The algorithm  has a meta-parameter $\mathcal{O}$, indicating the \textit{preference order}, which determines the order in which the properties of the referent are tested against the distractors. In our domain, for example, when \textit{color} is the most preferred property, the algorithm might return \textsc{blue}, if this property already excludes all distractors. When \textit{shape} is the preferred property and all distractors do \textit{not} share the shape \textsc{T} with the referent, \textsc{T} would be returned. Hence even when the referent and context are the same, different preference orders might lead to different expressions \citep{krahmer_is_2012}. We choose the preference order of \textit{color, shape and position} for the algorithm; we leave experimenting with other orders to future work.

\subsection{Data Generation}
\label{sec:data_generation}

The inputs $x_i$ for the task consist of two parts: the visual representation of the scene $v_i$ and a bounding box around the target piece $b_i$. For the automatic generation of these inputs we make use of \textit{symbolic board} representations $S_i=\{s_0,...,s_N\}$ where $N$ is the number of pieces on a board and $s_i$ is a tuple of \textit{color}, \textit{shape} and \textit{position} values e.g.  (\textsc{orange},\textsc{X},\textsc{top}). We define a mapping function $\mathcal{V}(S_i) \rightarrow v_i \in \mathbb{R}^{W \times H \times C}$ for rendering a board and sample uniformly from the symbols to select a target piece $b_i \sim S_i$  (for which we know the bounding box via $\mathcal{V}$). For simplicity, we use $b_i$ to refer to the target bounding box in the visual domain or the target piece in the symbolic domain respectively.

As the ground-truth expressions we define $y_i = \{w_0,...,w_T\}$ where $T$ is the length of the expression and $w_i$ is a word in the vocabulary. Again we make use of the symbolic piece representations (the same as above) to automatically generate the ground-truth by using the Incremental Algorithm. We apply the \textsc{ia} on the symbolic piece representations $S_i$ and the target symbol $b_i$ to select a set of property values $\mathcal{D}_i = \textsc{ia}(S_i,b_i)$ from the target $b_i$ (Algorithm~\ref{alg:ia}). These property values $\mathcal{D}_i$ are the shape, color or position values that are supposed to distinguish the target piece from other ones on the board. Finally, we define a mapping function $\mathcal{T}(\mathcal{D}_i) \rightarrow y_i$ to produce the ground-truth expression by filling the property values into pre-defined templates. %
The result of this process is a pairing of image and text, as you would find it for example in a captioning dataset \citep{johnson_densecap_2016}, albeit not collected from annotators but rather synthetically generated. 
In the following, we give more information on $\mathcal{V}$ and $\mathcal{T}$.

\subsection{$\mathcal{V}$: Rendering the Pentomino boards}

The symbolic piece representations in $S_i$ are rendered as visual inputs $v_i$. We implement the rendering function $\mathcal{V}(S_i)$ that paints the symbolic pieces according to their shape and color values with black borders onto a board of $30 \times 30$ same-sized tiles. 
This underlying grid is projected onto $224 \times 224$ pixels.
The exact tile coordinates of the pieces are determined by dividing the board into 9 distinct areas: one for each piece position value. To ensure that all pieces fit on the board, we allow maximal 2 pieces in a single area. We rotate and place the pieces one after the other into these areas by uniformly sampling tile coordinates that fall into the area that aligns to the piece position value. If two pieces collide during the placement, then we sample the coordinates again until they fit next to each other.%

\subsection{$\mathcal{T}$: Surface realization of \textsc{IA} outputs}

The \textsc{ia} returns properties $\mathcal{D}_i$ of a target piece that distinguish it from other pieces. This list of properties is then transformed into a natural language expression. %
We define a mapping function $\mathcal{T}$ that inserts the property values into one of 7 different templates (Appendix~\ref{appendix:templates}), for example ``Take the \textit{[color]} piece''. We call these templates \textit{expression types}. The mapping function selects the template based on the number of properties and the preference order: color, shape and then position. The word order in the templates is aligned with the preference order. We only use this order here to focus on the semantic correctness of the generation and leave mixing in additional variants like ``Take the \textit{[shape]} that is \textit{[color]} in \textit{[position]}'' to future work. 
Altogether the property values and the templates lead to a vocabulary of 38 words---an extremely small vocabulary, which however as we are not targeting lexical complexity here is not a problem. 

\begin{figure}[t]
    \begin{center}
        \includegraphics[width=0.35\textwidth]{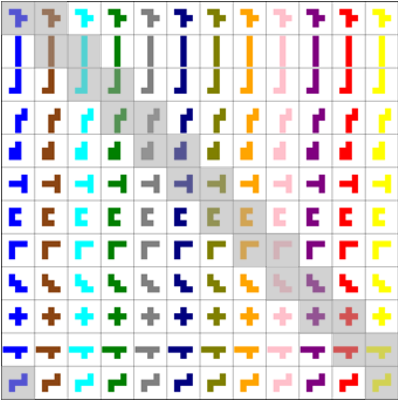}
    \end{center}
    \caption{The shape and color combinations in grey are never seen during training, but only during evaluation.}
    \label{fig:pento-color-holdouts}
    \vspace{-0.3cm}
\end{figure}

\subsection{Compositional Generalization}
\label{section:holdouts}

We make use of a synthetic dataset to guarantee the independence of properties and thus control, among other things, the compositionality of the learning task. %
There are $12$ conventional names for the shapes which are roughly inspired by visual similarity to letters like \textit{F}, \textit{T}, \textit{Y} etc. \citep{golomb_1996}. We sidestep the question of producing natural descriptions (``the one that looks like a boomerang'') for the shapes and assume that these letter names can be produced. Furthermore, the pieces can appear in one of $12$ different colors and the position can be approximated with $9$ different spatial expressions (Appendix~\ref{appendix:vocabulary}, \ref{appendix:colors}). The permutation of colors, shapes and positions leads to $12 \cdot 12 \cdot 9 = 1296$ symbolic pieces to choose from for the composition of boards and the selection of targets. Now we create the training data from only $|S_{\text{train}}|=840$ of the overall $1296$ possible piece symbols and leave the remaining ones as a ``holdout''. 
These holdout pieces are specifically used to test the models' generalisation along three different dimensions, as in the following.
\paragraph{Piece appearances \textit{(ho-color, 756 examples)}.} The target piece shapes are combined with new colors with respect to the training set (Figure~\ref{fig:pento-color-holdouts}). For each of the $12$ shapes we hold out $2$ colors (val,test). Then we generate for each shape-color combination one board for each position and expression type.
\paragraph{Piece positions \textit{(ho-pos, 840 examples)}.} In these examples the target pieces are shown at new positions with respect to the training set. For each of the pieces we hold out $2$ positions (val,test). Then we generate for the holdout combinations one board for each expression type.
\vspace{-3pt}

\paragraph{Expression types \textit{(ho-uts, 840 examples)}.} We test that expression types are not attributed to specific pieces and show them in new contexts that leads to new expressions types wrt.\ the training set. For each of the pieces we hold out $2$ expression types (val,test) and create  corresponding boards.

\begin{figure}[t]
    \begin{center}
        \includegraphics[width=0.45\textwidth]{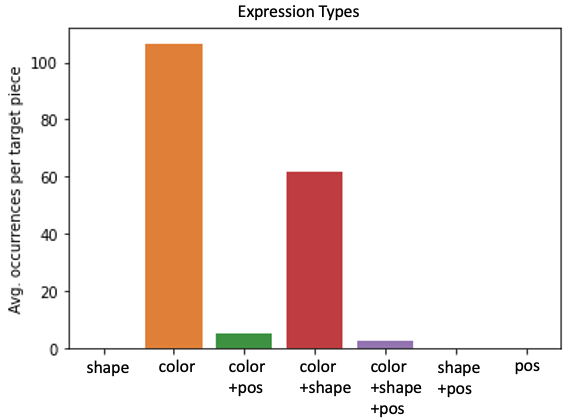}
    \end{center}
    \vspace{-3mm}
    \caption{In the \textsc{naive} dataset the synthesized examples highly prefer the generation of the two templates that only mention either the \textit{(color)} or \textit{(color,shape)} of a target piece. Some templates appear almost never [\textit{(shape)}, \textit{(shape,pos)}] or indeed never [\textit{(pos)}].}
    \label{fig:uts_random_sampling}
   \vspace{-0.3cm}
\end{figure}

\section{What data is necessary to learn the \textsc{ia}?}
\label{sec:dataset}

The learned models have to generate an expression with exactly those property values (not more and not less) that the \textsc{ia} would produce. We hypothesise that learning the iterative set logic process and the preference order (either implicitly or explicitly) from text and visual inputs alone constitutes a challenging task for them, especially because given 12 shapes, 12 colors and 9 (discrete) positions for a piece (minus the combinations we excluded for holdout), then there are already around 20 billion possibilities to produce a board with 4 pieces on it.

Thus we make use of the fact that the generation process is fully under our control and directly ask what kind of data distribution is necessary to learn this task. We experiment with two different dataset variants: The first variant (§\ref{section:naive}) relies on an unconstrained sampling of symbolic pieces for each board while the second variant (§\ref{section:didact}) is designed to be more informative through a curated selection process.

\subsection{\textsc{Naive}: Unconstrained Sampling}
\label{section:naive}

This process is meant to model the ``naive'' creation of a board by randomly sampling and placing pieces, as a person might do when setting up a board.
We create these examples by randomly filling boards with pieces: First, we decide on the number of pieces that go on the board by sampling $N$ from a uniform distribution over the integers 4--10. Then we sample uniformly with replacement $\{s_0,...,s_N\}$ from the $840$ symbolic pieces that are available for training. From the resulting symbolic board $S_i$ we choose one piece, again uniform random, as the target piece $b_{i_0}$. Finally, we generate the input pairing $(x_i,y_i)$ as described above. 

We add one further constraint: We re-use the visual board $v_i$ and pair it with 3 other target pieces $\{b_{i_1},b_{i_2},b_{i_3}\}$ chosen from $S_i$ without replacement, so that a model \textit{cannot} perform well by memorizing the $(x_i,y_i)$ pairings alone, because then there are 4 identical visual boards with different targets that lead to (most likely) different expressions.
This leads to 4 examples $(x_j,y_j)$ per visual board with $x_j=(v_{i_0},b_j)$. We repeat this procedure $42,000$ times which leads to $148,000$ training examples in total.
The quantitative evaluation shown in Figure~\ref{fig:uts_random_sampling} reveals that here a model is most of the times confronted with expressions that only mention the \textit{color} value or the \textit{color and shape} of the target piece. 
The orange bar indicates that there are on average 100 examples (board and target) for each of the possible 840 target pieces where the color alone uniquely identifies the piece. So for around 84K samples in this dataset, a sentence like ``Take the [blue, red, green,...] piece'' would be correct. 

\begin{figure}[t]
    \begin{center}
        \includegraphics[width=0.44\textwidth]{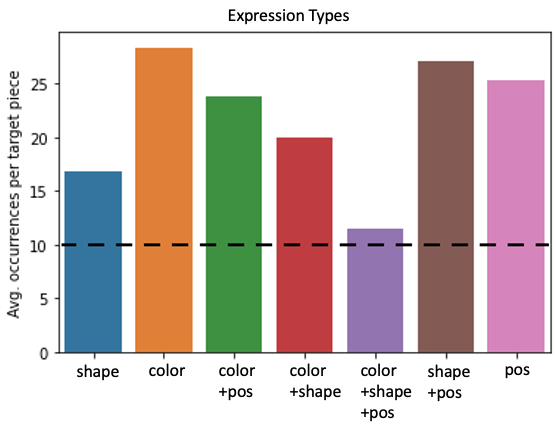}
    \end{center}
    \vspace{-3.5mm}
    \caption{The \textsc{didact} dataset is fairly balanced with respect to the expressions types.
    The figure shows the average count of expression types for each target piece after generating 
    with \textsc{etos}. The dashed line indicate the distribution without the extra target selection.}
    \label{fig:uts_etos_sampling}
    \vspace{-0.3cm}
\end{figure}

\subsection{\textsc{didact}: Expression Oriented Sampling}
\label{section:didact}

The goal of the alternative sampling process is to ensure that examples of all \textit{output} types are represented in the dataset, in a balanced way. We assume that this results in a more ``didactic'' dataset from which the underlying relation between input and desired output can more easily be induced. 
The idea is to directly choose the distractors of a target piece in such a way that the wanted expression type has to be produced. For example, when the target piece is \textsc{(orange, X, top)} and the expression type is supposed to be \textit{Take the [shape]}, then we construct a set of distractors where some share color and position, but none is of shape \textsc{X}. We call this approach \textit{expression type oriented sampling} (\textsc{etos}) (details in  Appendix~\ref{appendix:etos_rules}).
This method allows us to confront the learner with all the possible expressions about the same amount of times. Thus each target piece is seen on $50$ different boards resulting in $840 \cdot 50 = 42,000$ boards (Table~\ref{table:data_splits_symbolic}). 

Yet again we avoid that $(x_i,y_i)$ pairs can be simply memorized and select as before 3 other pieces as the targets $\{b_{i_1},b_{i_2},b_{i_3}\}$  which leads to $148,000$ examples in total.
The consequences of the extra target selection within this method are twofold: Firstly, the distribution is a bit shifted towards the \textsc{naive} approach as shown in Figure~\ref{fig:uts_etos_sampling} because we randomly select the target, and more importantly there might be now expressions produced that were actually intended for the holdout \textit{(ho-uts)}. We remove such ``unintended'' examples from the training set so that there are $128,526$ examples for training (Table~\ref{table:data_splits_training}).
Whereby the guarantees we can make for this ``\textsc{didact}ic'' dataset are that:

\vspace{-3pt}
\paragraph{Target pieces appear with different distractors.} Each target piece symbol for training appears on average in $153$ contexts as a target.

\vspace{-4pt}
\paragraph{On the same board occur different target pieces.} We choose 3 additional pieces as targets apart from the one for which the board was initially intended.

\vspace{-4pt}
\paragraph{Target pieces appear also on other boards.} Each symbolic piece appears on average in $1,075$ contexts, which is more often than as a target.

\begin{table}[t]
\small
\begin{tabular}{|l|r|r|r|r|}
\hline
\multicolumn{1}{|r|}{}                    & \multicolumn{1}{l|}{} & \multicolumn{1}{l|}{} & \multicolumn{1}{l|}{Bords}  & \multicolumn{1}{l|}{Boards} \\
Dataset / Num. of                       & \multicolumn{1}{l|}{TPS}    & \multicolumn{1}{l|}{pET}   & \multicolumn{1}{l|}{per pET} & \multicolumn{1}{l|}{Total}  \\ \hline
\textsc{naive}  & 840                      & 7                       & -                         & 42,000                       \\ \hline
\textsc{didact} & 840                      & 5                       & 10                          & 42,000                       \\ \hline
ho-uts val                                & 840                      & 1                       & 1                           & 840                         \\ \hline
ho-uts test                               & 840                      & 1                       & 1                           & 840                         \\ \hline
ho-color val                              & 108                      & 7                       & 1                           & 756                         \\ \hline
ho-color test                             & 108                      & 7                       & 1                           & 756                         \\ \hline
ho-pos val                                & 120                      & 7                       & 1                           & 840                         \\ \hline
ho-pos test                               & 120                      & 7                       & 1                           & 840                         \\ \hline
\end{tabular}
\vspace{-.2cm}
\caption{The number of target piece symbols (TPS) and possible expression types (pET) per TPS for the \textsc{naive}, \textsc{didact} and holdout datasets. Although the number of the resulting \textsc{naive} and \textsc{didact} boards is with $42,000$ the same, the boards are generated with different techniques: either with random uniform sampling (\textsc{naive}; the number of boards per pET and TPS is not controlled for) or expression type oriented sampling (\textsc{didact}; 10 boards for each pET and TPS). 
}
\label{table:data_splits_symbolic}
\vspace{-.3cm}
\end{table}

\begin{table}[t]
\small
\begin{tabular}{|l|r|r|}
\hline
              & \multicolumn{1}{l|}{\textsc{naive} dataset} & \multicolumn{1}{l|}{\textsc{didact} dataset} \\ \hline
Number of Boards  & 42,000                               & 42,000                                \\ \hline
TPS per Board & 4                                   & 4                                    \\ \hline
Number of Samples & 168,000                              & 168,000                               \\ \hline
Validation    & 10,000                               & 10,000                                \\ \hline
Testing       & 10,000                               & 10,000                                \\ \hline
Training      & 148,000                              & 148,000                               \\ \hline
Filtered      & 128,526                              & -                               \\ \hline
\end{tabular}
\vspace{-.2cm}
\caption{The number of samples for each dataset and training split. For each board we chose 4 target piece symbols  (TPS) (incl. the originally intended target piece in the \textsc{didact} dataset) resulting into $168,000$ samples for both datasets. From this overall samples we choose $10,000$ for validation and testing. In addition, we exclude the training samples of the \textsc{didact} for which an expression is to be produced that should reserved for the \textit{ho-uts} testing. This is not done for the \textsc{naive} dataset because here we train on possibly all expression types.}
\label{table:data_splits_training}
\vspace{-.3cm}
\end{table}

\section{Learning the Incremental Algorithm}
\label{sec:learning}

Our goal in producing the collection of scenes was to ensure that a model $f$ must indeed be based on features of the $x_i$ that we care about (that is, which figure in the desired capability), namely the need to indeed compare the perceivable target piece and distractor properties. The \textsc{ia} (§\ref{sec:ia}) achieves this by a hard-coded loop structure over symbols which (a) compares the objects (b) sticks to a preference order (c) preemptively stops when all distractors are excluded and (d) outputs the uniquely identifying properties (or all properties in ambiguous cases).

In the following, we present our neural models (§\ref{sec:models}) and the conducted experiments (§\ref{sec:experiments}) to test if neural language generators are indeed able to acquire such a ``programatic'' capability by the simple task definition of producing expressions from visual inputs. The generation models $f$ will be trained on the basis of $(x_i, y_i)$ pairs only.
We train two common network architectures for this task of which one is an LSTM-based approach to REG proposed by \citet{refcocog} for natural scenes and the other is a transformer \citep{vaswani_attention_2017}. In addition, we propose a variant for processing the inputs along with a simple classifier-based baseline.

\subsection{Models}
\label{sec:models}

\begin{figure}[t]
    \begin{center}
        \includegraphics[width=0.45\textwidth]{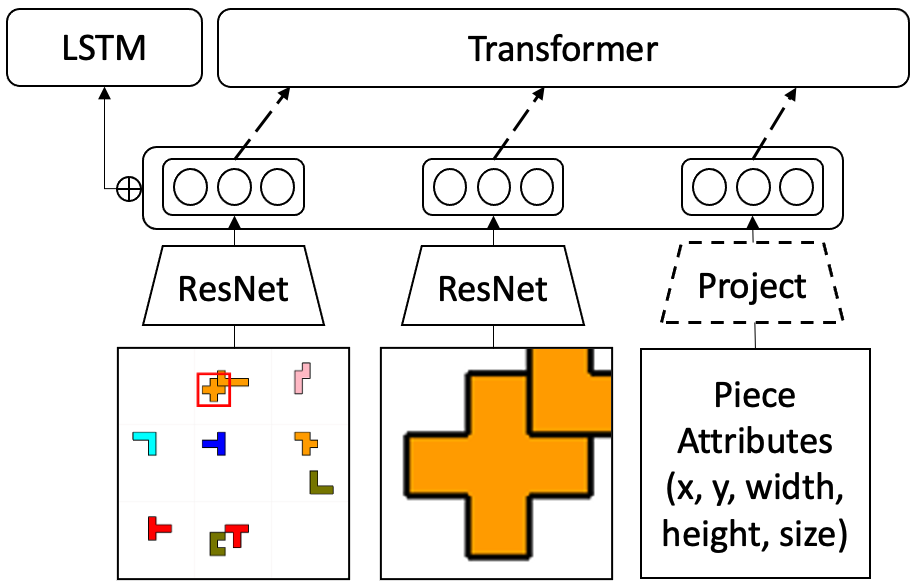}
    \end{center}
    \caption{The encoding of the visual scene, target piece and its attributes as proposed by \citet{refcocog}. The dashed lines indicate the transformers' information flow.}
    \label{fig:example_seq_input}
    \vspace{-0.3cm}
\end{figure}

\paragraph{LSTM.} \citet{refcocog}, who present a model of REG in natural scene images, embed the scenes and the referent within them with a pre-trained VGG \citep{liu_very_2015}. We follow their procedure but use the $512$ dimensional embeddings after global average pooling of a ResNet-34  \citep{he_deep_2016} and fine-tune all of its layers because our images look very different to the ones from the pre-training on ImageNet \citep{deng_imagenet_2009}. 
We cut out the target piece 
using the bounding box information. Then the piece snippet is dilated with 5 context pixels %
and up-scaled to the size of the board image. We additionally randomly shift the snippet by 0-5\% of the pixels in either direction horizontally or vertically (fill-color is white). The target piece and board image embeddings are then concatenated together with five location and size features of the target. The resulting 1029-dimensional feature vector is fed to an LSTM at each time step to condition the language production (using greedy decoding). We reduce the word embedding dims to $512$ because our vocabulary is very small and apply an Adam optimizer \citep{Kingma2015AdamAM}.

\paragraph{Transformer.} For comparison with \citet{refcocog} we resize, augment and encode the target piece and visual board with a ResNet-34 in the same way as described before. Then the image embeddings are fed into the transformer \citep{vaswani_attention_2017}  individually (not concatenated) as ``visual words'' together with the target piece attributes embedding as shown in Figure~\ref{fig:example_seq_input} to compute an intermediate representation of the inputs altogether. This ``memory'' embedding is then fed into the decoder to generate the RE using masked self-attention as in other machine translation tasks. For the variable length expressions we use a padding symbol and ignore prediction at padded positions during loss computation. We reduced the original capacity of the model to avoid overfitting and applied a learning rate scheduling strategy as described by \citet{vaswani_attention_2017}, using an AdamW optimizer \citep{Loshchilov2019DecoupledWD}.

\begin{figure}[t]
    \begin{center}
        \includegraphics[width=0.45\textwidth]{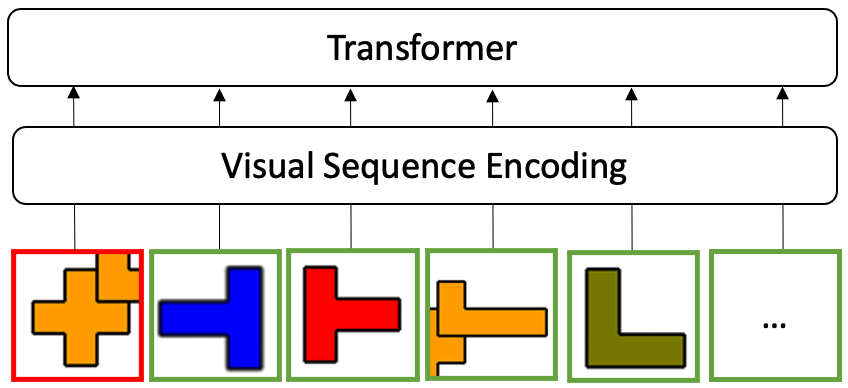}
    \end{center}
    \caption{An exemplary visual input sequence for the Transformer+\textsc{vse} model. The target is highlighted with a red and the distractors with a green border.}
    \label{fig:transformer_vse}
    \vspace{-0.3cm}
\end{figure}

\paragraph{Transformer+\textsc{vse}.} We assume that the transformer should be particularly capable of
generating \textsc{ia}-like expressions because self-attention might allow it to learn the required piece-wise comparison operation. The self-attention mechanism has already been proven powerful for other image-related tasks \cite{Li2020OscarOA, Zhang2021VinVLRV,jaegle_perceiver_2021}. Therefore we follow \citet{tan_lxmert_2019} and implement a \textit{visual sequence encoding} (\textsc{vse}) mechanism. For this we cut out each piece on the board to produce a sequence of piece snippets as shown in Figure~\ref{fig:transformer_vse} and project the visual features onto the models' input dimensions $\hat{f_j}$ and add region embeddings $\hat{p_j}$ to them that contain the snippets size and location information:\footnote{$b_F,b_P$ are bias terms of the linear projections}

\vspace{-0.5cm}
\begin{align}
   \hat{f_j} &= \text{LayerNorm} (W_F f_j + b_F) \\
   \hat{p_j} &= \text{LayerNorm} (W_P p_j + b_P) \\
    v_j &= ( \hat{f_j} + \hat{p_j} + \hat{t_j} ) / 3 \label{eq:sum}
\vspace{-0.1cm}
\end{align}

To let a model distinguish between target and distractor ``words'' in the input sequence we add a type embedding $\hat{t_j}$, similar to word embeddings, and normalize. Furthermore, we have a variable amount of pieces on the board (between 4 and 10), but a transformer model assumes a fixed-size input sequence (per batch, during training). Thus we indicate ``padding'' pieces with a padding index in the sequence as implemented in PyTorch \citep{pytorch_2019} and use images with all zeros for them.

\begin{table*}[t]
\centering
    \begin{small}
    \begin{tabular}{|l|c||c|c|c|c||c|c|c|c|c|}
        \hline
         Model      & Data  & \multicolumn{4}{c||}{BLEU@1 (in \%) $\uparrow$} & \multicolumn{4}{c|}{Sentence-wise Acc. (in \%) $\uparrow$} \\
                    &       & in-dist. & ho-color & ho-pos & ho-uts & in-dist. & ho-color & ho-pos & ho-uts \\
        \hline
         LSTM \cite{refcocog}     & \textsc{naive} & 38   & 33   & 33 & 34        & 24 & 17     & 18 & 18 \\
         LSTM  \cite{refcocog}     & \textsc{didact}  & 64   & 62   & 62 & 51        & 31 & 24     & 24 & 4 \\
         \hline
         Transformer & \textsc{naive} & 27   & 23   & 23 & 23    & 21 & 14     & 14 & 15 \\
         Transformer & \textsc{didact} & 79   & 77   & 76 & 76     & 53 & 53     & 51 & 33 \\
         \hline
         Transformer+\textsc{vse}   & \textsc{naive} & 59   & 53   & 57 & 54     & 29 & 22     & 22 & 25 \\
         Transformer+\textsc{vse}   & \textsc{didact}  & \textbf{97} & \textbf{97} & \textbf{97} & \textbf{97} & \textbf{91} & \textbf{91} & \textbf{91} & \textbf{92}\\
         \hline
         Classifier+\textsc{vse} & \textsc{naive}  & 32   & 25   & 28 & 28     & 27 & 15     & 19 & 22 \\
         Classifier+\textsc{vse} & \textsc{didact}   & 91   & 79   & 77 & 60     & 76 & 40     & 44 & 14 \\
        \hline
    \end{tabular}
    \end{small}
    \caption{The match rates for unigrams (BLEU@1) and whole sentences (SentA) on the test splits. The \textit{in-dist.} samples are from the \textsc{didact} dataset because these are more balanced with respect to the expression types.}
    \label{tab:results}
    \vspace{-0.3cm}
\end{table*}

\paragraph{Classifier+\textsc{vse}.} The representations of the \textsc{vse} might already capture enough information to perform the task. Therefore we test this assumption by training a simple linear sentence classifier just on top of the concatenated embeddings. The classifier has to predict the correct sentence out of the 1,689 possible ones. This framing is similar to that often used in visual question answering \citep{Hudson2019GQAAN}, where the possible answers are framed as classes in a classification task.

\subsection{Metrics}
\label{sec:evaluation}

We use the well known and commonly reported precision-based BLEU@1 metric for evaluation because this is simple metric for word matching when having only a single reference.
In addition, we compute the \textit{sentence-wise accuracy} (SentA) that indicates how often a prediction does exactly match the single reference so that the order of the words matters. As an example in Appendix~\ref{appendix:qualitative_analysis} the model erroneously produces ``Take the i top in the top left''. We ignore the starting words ``Take the'' for the evaluation when they occur in both the prediction and the ground-truth, because then they are  uninformative about the real performance.

\subsection{Experiments}
\label{sec:experiments}

We perform separate training runs on both a \textsc{naive} (§\ref{section:naive}) and \textsc{didact}  (§\ref{section:didact}) dataset for a maximum of 100 epochs 
and perform 10 validation runs during an epoch. Over all validation runs we save the three best performing models with respect to the BLEU@1 score using greedy decoding. We stop the training when the model does not improve anymore after 20 validation runs. For evaluation we choose the model with more epochs if the scores are the same. The training objective is to minimize the cross-entropy between the predicted and the ground-truth expression given by the Incremental Algorithm (\textsc{ia}). 

\section{Results and Discussion}
\label{sec:results}

\paragraph{\textsc{naive} versus \textsc{didact}.} The results in Table~\ref{tab:results} show that even the worst performing model trained on the \textsc{didact} dataset (LSTM 31\% in-dist) \textit{is still performing better} than the best performing model trained on the \textsc{naive} dataset (Transformer+\textsc{vse} 29\% in-dist) over all SentA scores (except ho-uts). This indicates that a well controlled data generation procedure is essential to perform well on this task, or conversely, that none of the learning algorithms can guess at the underlying minimality constraint from the unconstrained data alone.
The SentA scores for the \textsc{naive}-based models indicate that these often perform only about by chance (picking 1 of 7 templates leads to a score of 14\%) on the compositional splits (highest 25\% and avg.\ 18\%). These splits contain all expression types in equal amounts and we find that these \textsc{naive} models tend to produce only a few expression types. 

\paragraph{Input triplets versus \textsc{vse}.} The results show that for both datasets a significant increase in performance is achieved by using \textsc{vse} which includes a vision detection step. For the Transformer+\textsc{vse} model the BLEU@1 scores double from 27\% to 59\% on the in-distribution test data. The simple Classifier+\textsc{vse} model performs similarly well as the other models \textit{without \textsc{vse}}. This is reasonable because with \textsc{vse} the visual encoder must \textit{not} operate on two different image resolutions anymore: one for the (up-scaled) target piece and one for the whole context image. The \textsc{vse} detection step ``frees'' capacities that would be necessary to correctly identify the content of the context image.

\paragraph{Classifier+\textsc{vse} versus others.} Almost all models struggle to perform well on both the compositional (<54\% SentA) and the in-dist. test data (<77\% SentA). Thus the Classifier+\textsc{vse} establishes a relative high baseline on most of the test sets (76\%/40\%/44\%) but only performs about by chance (14\%) at the \textit{ho-uts} data (which contains unseen expression types). The Transformer+\textsc{vse} model is the only one that exceeds the high Classifier+\textsc{vse} baseline by achieving almost perfect scores (91\% SentA) over all categories when trained on the \textsc{didact} data. 

\paragraph{Effect of individual input features.} We perform an ablation study to measure the impact of particular input features on the SentA scores. We do so by replacing the individual parts of the visual sequence encoding of our best model with noise sampled from a standard gaussian. We see that the visual embeddings are essential to generate the correct referring expression as the sentence-wise accuracy drops to 1\% (Table~\ref{tab:ablation}). A similar performance drop is seen for the type embeddings where the accuracy is only 1-2\%. A different impact is measured, when the region embeddings are replaced with random noise; here the accuracy is still around 40-44\%. This is reasonable, because in only 4 of the 7 expression templates, the position (and therefore the region embeddings) are relevant.

\begin{figure}[t]
     \centering
     \includegraphics[width=0.48\textwidth]{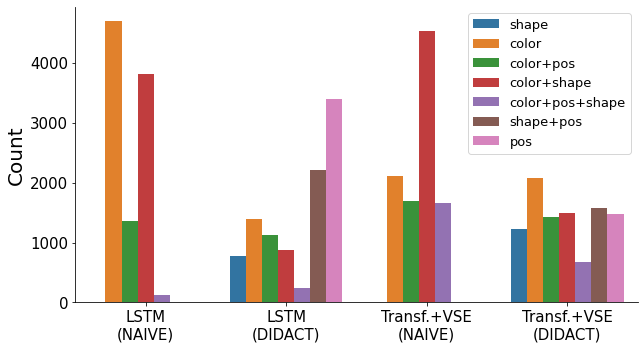}
     \vspace{-6.0mm}
     \caption{Produced expression types on the \textit{in-dist}. split.}
     \label{fig:ut_preds_data}
     \vspace{-2.0mm}
\end{figure}

\paragraph{Effect of \textsc{didact}ic training.} We have a closer look on the tendencies of the models to produce certain expression types on the test data. For this we applied a parser to the predicted expressions of the models and counted the expression type occurrences. This provides insights, if a model tends to ``overfit'' on specific expression types. For example as the surface structure of the \textit{color} expression types is seen in majority of cases during training, a model might simply try to produce \textit{Take the [color] piece} and insert the referent color. We do not check for the correctness of the produced expressions here.
The measures show that the LSTM model trained on the \textsc{naive} dataset has converged on a behavior that produces in the majority of cases the \textit{color} or \textit{color+shape} expression type (Figure~\ref{fig:ut_preds_data}). This is reasonable as this is the majority class in the random sampling data. Only the \textsc{didact} dataset let's them pick up on other expression types more regularly. The Transformer+\textsc{vse} produces on the \textsc{didact} test dataset rather balanced amounts of expression types (as these are given in the test data).

\section{Conclusion and Future Work}
\label{sec:conclusion}

In this work we presented the diagnostic dataset \datasetname\ to study the question whether neural models can learn the RE production strategy of the Incremental Algorithm (\textsc{ia}). A symbolic algorithm that is motivated by the appeal to the hypothesises capability of ``elimination of distractors'' (through the application of Gricean maxims).

We found through the better control on scene complexity that an unconstrained sampling method (\textsc{naive}) does not provide enough information for a neural model to pick up on the underlying regularity and to exhibit the desired capability, while an output oriented sampling process (\textsc{didact}) does. This indicates that the generalizability in this task and domain is not given by the capabilities of the learner alone but is strongly %
determined by the learning examples.
We evaluated a classic LSTM-based model and a modern transformer (that have to process two different image resolutions) and observed that these still struggle even on the more informative dataset (\textsc{didact}). We proposed a modification of the input processing that comes with a detection step (\textsc{vse}) and observed that this leads to a strong baseline and allows the transformer  to converge. This indicates that object detection is an essential requirement to perform well on this task.

In future work we want to evaluate more models on our diagnostic dataset to find potential weaknesses. An interesting question is whether a PLM \citep{gpt3} might have picked up such Gricean constraints and would be able to recognise their desirability from being prompted with only a few examples. We also plan to explore to what extent our best model is applicable to more realistic settings following \textit{Sim-to-Real} approaches \citep{sim2real}.

\begin{table}[t]
    \begin{small}
    \centering
    \begin{tabular}{|l|c|c|c|c|}
        \hline
                        & \multicolumn{4}{c|}{Sentence-wise Acc. (in \%) $\uparrow$} \\
        Transformer+\textsc{vse}   &  in-d. & ho-color & ho-pos & ho-uts \\
        \hline
         w/o visual emb. &     1 &        1 &      1 &      1 \\
         w/o type emb.   &     2 &        1 &      2 &      2 \\ 
         w/o region emb. &    44 &       42 &     41 &     40 \\ 
         full model       &    91 &       91 &     91 &     92 \\ 
         \hline
    \end{tabular}
    \vspace{-2.0mm}
    \caption{The ablation study performed on the test dataset shows that the visual and type embeddings contain the crucial information for the RE generation.}
    \label{tab:ablation}
    \end{small}
\end{table}

\clearpage

\section*{Limitations}

\paragraph{Limits on visual variability and naturalness.} The Pentomino domain can only serve as an abstraction for referring expression generations in visual domains. The amount of objects is limited to 12 different shapes and the number of colors is reduced to 12 as well. The positions are chosen to be discrete and absolute while real-world references might include spatial relations which we leave for further work. Furthermore, the pieces show no texture or naturalness, but are drawn with a solid color fill and a simple black border. Various lightning conditions that might impact a vision detection system are avoided. We left the evaluation of the proposed models on more realistic dataset for further work.

\paragraph{Limits on variability of the referring expressions.} We only explored expressions that are generate by the Incremental Algorithm with one fix preference order of color, shape and position although we are aware of the fact that preference order might vary between subjects \citep{krahmer_is_2012}. Moreover, we choose a fix property value order (color is mentioned before shape is mentioned before position) for the realisation of the template's surface structure and left the exploration for a higher variability to further work.

\paragraph{Limits possible claims about human capabilities.} As this work is on synthetic dataset created by an algorithm, any claims about human capabilities, and about a model's ability to acquire those, are only made indirectly, via the quality of the original algorithm. 

\section*{Acknowledgements} We want to thank the anonymous reviewers for their comments. This work was funded by the \textit{Deutsche Forschungsgemeinschaft} (DFG, German Research Foundation) – 423217434 (``RECOLAGE'') grant.

\bibliographystyle{acl_natbib}
\bibliography{custom}

\appendix
\label{sec:appendix}

\section{Experiment details}
\label{appendix:details}

We trained each of our models on a single GeForce GTX 1080 Ti (11GB).

\subsection{The vocabulary}
\label{appendix:vocabulary}

The vocabulary includes the following 38 words:

\begin{itemize}
    \item 12 shapes: F, I, L, N, P, T, U, V, W, X, Y, Z
    \item 12 colors: red, orange, yellow, green, blue, cyan, purple, brown, grey, pink, olive green, navy blue
    \item 6 position words: left, right, top, bottom, center (which are combined to e.g., right center or top left)
    \item 4 template words: Take, the, piece, at
    \item 4 special words: <s>, <e>, <pad>, <unk>
\end{itemize}

\subsection{The piece colors (RGB-values)}
\label{appendix:colors}

\begin{table}[h]
    \centering
    \begin{tabular}{|l|l|l|}
    \hline
    Name & HEX & RGB \\
    \hline
 red & \#ff0000 & (255, 0, 0) \\
 orange & \#ffa500 & (255, 165, 0) \\
 yellow & \#ffff00 & (255, 255, 0) \\
 green & \#008000 & (0, 128, 0) \\
 blue & \#0000ff & (0, 0, 255) \\
 cyan & \#00ffff & (0, 255, 255) \\
 purple & \#800080 & (128, 0, 128) \\
 brown & \#8b4513 & (139, 69, 19) \\
 grey & \#808080 & (128, 128, 128) \\
 pink & \#ffc0cb & (255, 192, 203) \\
 olive green & \#808000 & (128, 128, 0) \\
 navy blue & \#000080 & (0, 0, 128) \\
    \hline
    \end{tabular}
    \caption{The colors for the Pentomino pieces. We also have 2 two-word colors: olive green and navy blue.}
    \label{tab:my_label}
    \vspace{-0.3cm}
\end{table}

\subsection{Uniform distribution of piece properties}

\begin{figure}[h]
    \begin{center}
        \includegraphics[width=\linewidth]{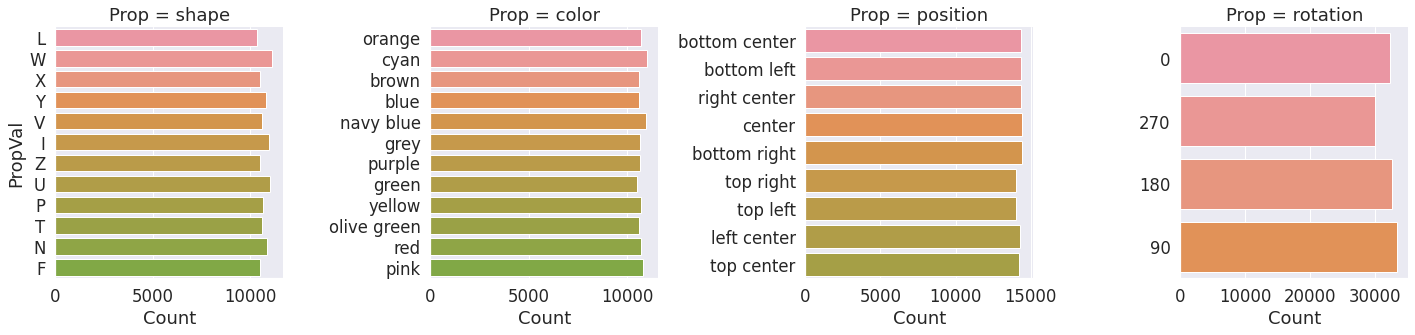}
    \end{center}
    \caption{The occurrences of property values of target pieces in the \textsc{didact} training data are almost uniform.}
    \label{fig:dist_property_values}
    \vspace{-0.3cm}
\end{figure}

Our expression type oriented sampling strategy achieves an almost uniform distribution of piece color, shapes and positions (even rotations) as shown in Figure~\ref{fig:dist_property_values}. We ignore the rotation property, but apply it to make the task harder. The model has to become invariant to the rotation.

\section{Data Generation}

\subsection{\textsc{didact} dataset generation details}
\label{appendix:etos_rules}

To construct this data, we iterate over all possible training symbols in $S_{train}$ and set them as the target piece $b_i$ directly (Table~\ref{table:data_splits_symbolic}). Then we sample a symbolic board $S_i$ (that includes the target) from the set of possible symbolic boards that lead to a wanted expression type $u_i$. For this we define the generator function $\mathcal{G}(u_i,b_i)$ that finds all possible symbolic boards that will be mapped by $\mathcal{T}$ so that $\{ S_{u_i} |\mathcal{T}(\textsc{ia}(S_{u_i},b_i)) \in \mathcal{Y}(u_i) \}$ where $\mathcal{Y}(u_i)$ is the collection of expressions that are represented by the template $u_i$, for example ``Take the \textit{[red, blue, green,...]} piece''. In a sense $\mathcal{G}$ is the inverse of $\mathcal{T}$.

This method allows us to confront the learner with all the possible expressions about the same amount of times. We perform the example generation 10 times for each target piece and the according 5 training expression types (see §\ref{section:holdouts} for holdouts). Finally, we generate the input pairing $(x_i,y_i)$ as described in §\ref{sec:task}.
Thus each target piece is seen on $50$ different boards resulting in $840 \cdot 50 = 42,000$ boards. Yet again we avoid that $(x_i,y_i)$ pairs can be learnt by heart and select as before 3 other pieces as the targets $\{b_{i_1},b_{i_2},b_{i_3}\}$  which leads to $148,000$ samples in total of which we filter the unintended ones (Table~\ref{table:data_splits_training}).

\subsection{Holdout generation details}

For the \textit{ho-color} and \textit{ho-pos} splits we additionally allow to choose distractors from the $840$ symbolic pieces of the training split, because otherwise the distractor set of possible piece might become empty e.g. for the \textit{ho-pos} split we have  the target pieces only on a subset of possible positions, but need to place distractors in additional positions to produce all expression types.

\section{Expression Types}
\label{appendix:templates}

There are 3 expression types that are used when only a single property value of the target piece is returned by the Incremental Algorithm (\textsc{ia}): 
\begin{itemize}
    \itemsep0em
    \item \textit{Take the [color] piece}
    \item \textit{Take the [shape]}
    \item \textit{Take the piece at [position]}
\end{itemize}
Then there are 3 expression types that are selected when two properties are returned:
\begin{itemize}
    \itemsep0em
    \item \textit{Take the [color] [shape]}
    \item \textit{Take the [color] piece at [position]}
    \item \textit{Take the [shape] at [position]}
\end{itemize}
And finally there is one expression type that lists all property values to identify a target piece:
\begin{itemize}
    \itemsep0em
    \item \textit{Take the [color] [shape] at [position]}
\end{itemize}

In the following we exemplify the generated boards for each of the expression types.

\subsection{Take the [color] piece}
\begin{figure}[h]
    \begin{center}
        \includegraphics[width=0.35\textwidth]{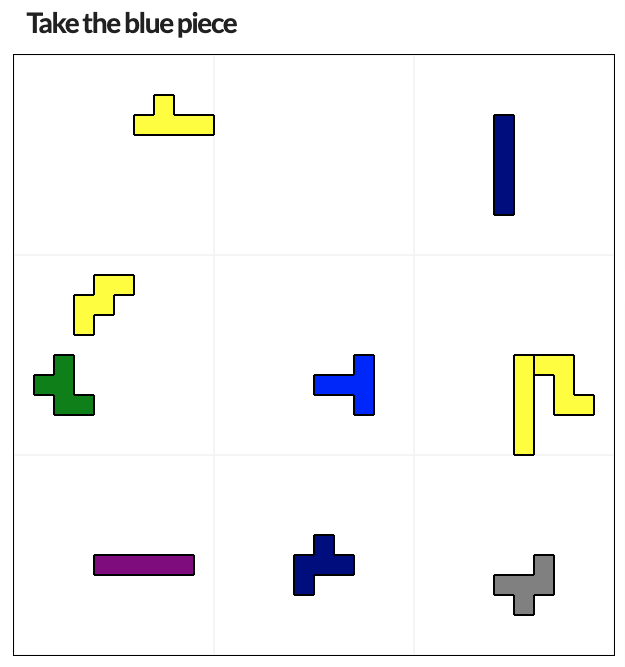}
    \end{center}
    \vspace{-0.0cm}
    \caption{A sample board with the target piece \texttt{(T,blue,center)} for this expression type.}
    \label{fig:ut_1}
    \vspace{-0.3cm}
\end{figure}

Mention the \textbf{color} excludes all. We add distractors with any shape or position, but a different color.

\subsection{Take the [shape]}
\begin{figure}[h]
    \begin{center}
        \includegraphics[width=0.35\textwidth]{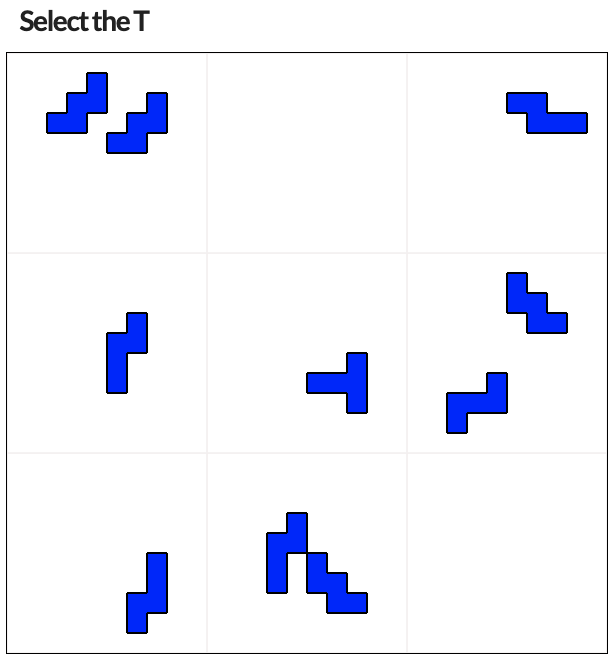}
    \end{center}
    \vspace{-0.0cm}
    \caption{A sample board with the target piece \texttt{(T,blue,center)} for this expression type.}
    \label{fig:ut_2}
    \vspace{-0.3cm}
\end{figure}

Mention the \sout{color} does not exclude any. Mention the \textbf{shape} excludes all. We add distractors with the same color, but different shape and at any position.

\newpage

\subsection{Take the piece at [position]}

\begin{figure}[h]
    \begin{center}
        \includegraphics[width=0.35\textwidth]{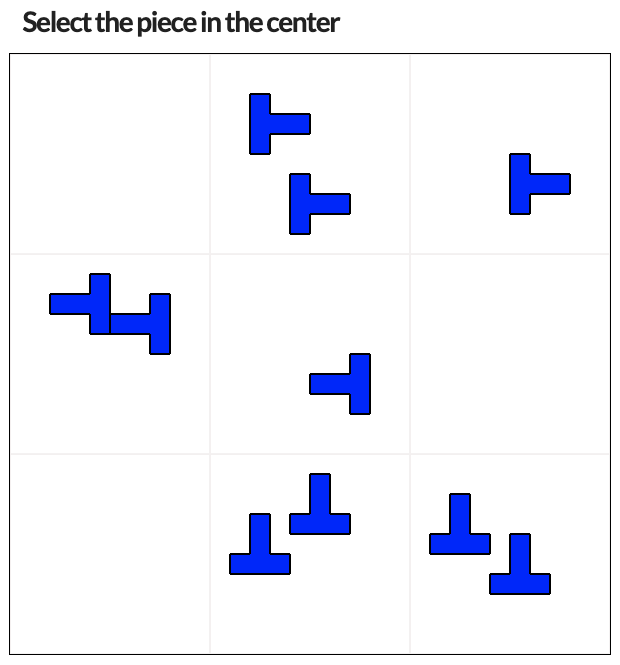}
    \end{center}
    \vspace{-0.0cm}
    \caption{A sample board with the target piece \texttt{(T,blue,center)} for this expression type.}
    \label{fig:ut_3}
    \vspace{-0.3cm}
\end{figure}

Mention the \sout{color} does not exclude any. Mention the \sout{shape} does not exclude any. Mention the \textbf{position} excludes all. We add distractors with the same color and shape, but at a different position.

\vspace{0.95cm}

\subsubsection{Take the [color] [shape]}

\begin{figure}[h]
    \begin{center}
        \includegraphics[width=0.35\textwidth]{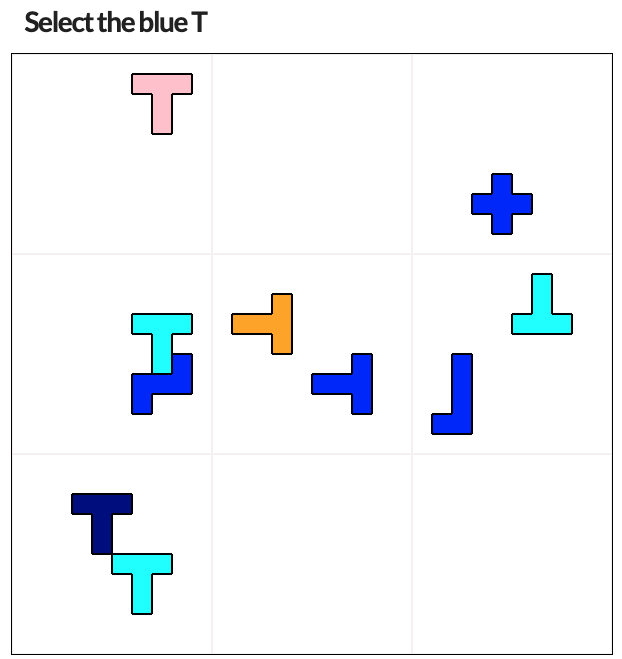}
    \end{center}
    \vspace{-0.0cm}
    \caption{A sample board with the target piece \texttt{(T,blue,center)} for this expression type.}
    \label{fig:ut_4}
    \vspace{-0.3cm}
\end{figure}

Mention the \textbf{color} excludes some, but not all. Mention the \textbf{shape} excludes the rest. We add some distractors with the same color (but different shape) and some distractors with the same shape (but different color) at any position.

\newpage

\subsubsection{Take the [color] piece at [position]}

\begin{figure}[h]
    \begin{center}
        \includegraphics[width=0.35\textwidth]{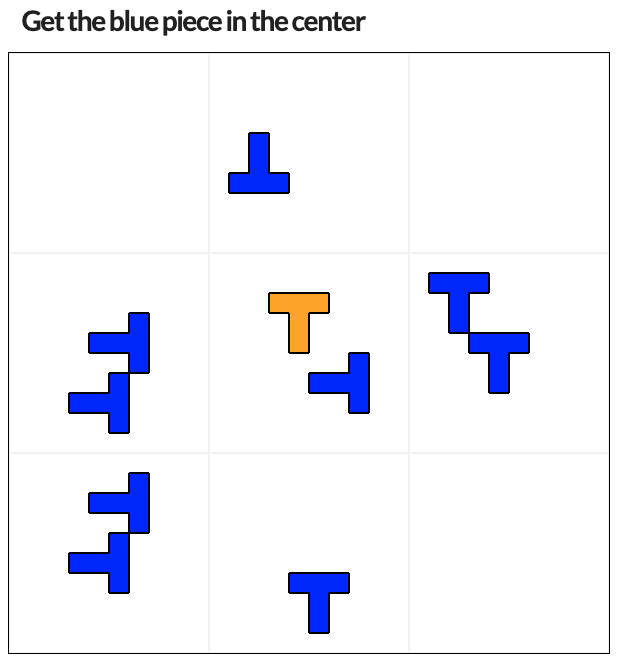}
    \end{center}
    \vspace{-0.0cm}
    \caption{A sample board with the target piece \texttt{(T,blue,center)} for this expression type.}
    \label{fig:ut_5}
    \vspace{-0.3cm}
\end{figure}

Mention the \textbf{color} excludes some, but not all. Mention the \sout{shape} does not exclude any. Mention the \textbf{position} excludes the rest. We add some distractors with the same color (but different position) and some with the same position (but different color) and the same shape.

\subsubsection{Take the [shape] at [position]}

\begin{figure}[h]
    \begin{center}
        \includegraphics[width=0.35\textwidth]{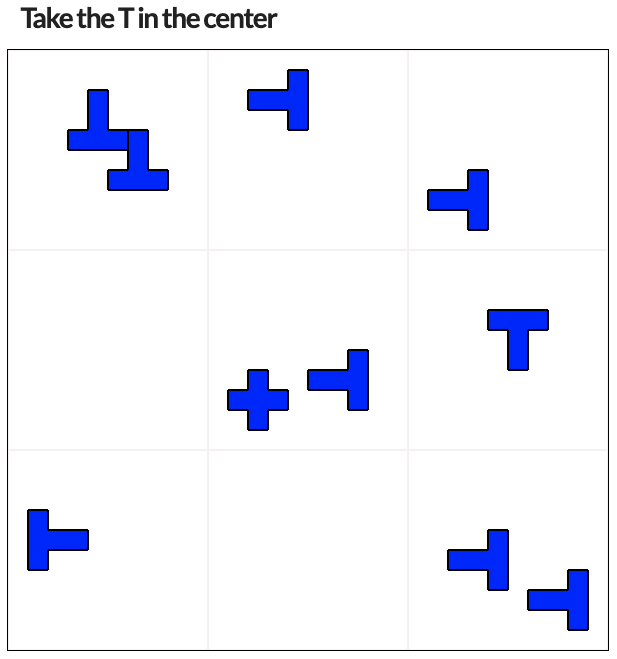}
    \end{center}
    \vspace{-0.0cm}
    \caption{A sample board with the target piece \texttt{(T,blue,center)} for this expression type.}
    \label{fig:ut_6}
    \vspace{-0.3cm}
\end{figure}

Mention the \sout{color} does not exclude any. Mention the \textbf{shape} excludes some, but not all. Mention the \textbf{position} excludes the rest. We add distractors with the same color and some with the same shape (but different position) and some with the same position (but different shape).

\newpage

\subsubsection{Take the [color] [shape] at [position]}

\begin{figure}[h]
    \begin{center}
        \includegraphics[width=0.35\textwidth]{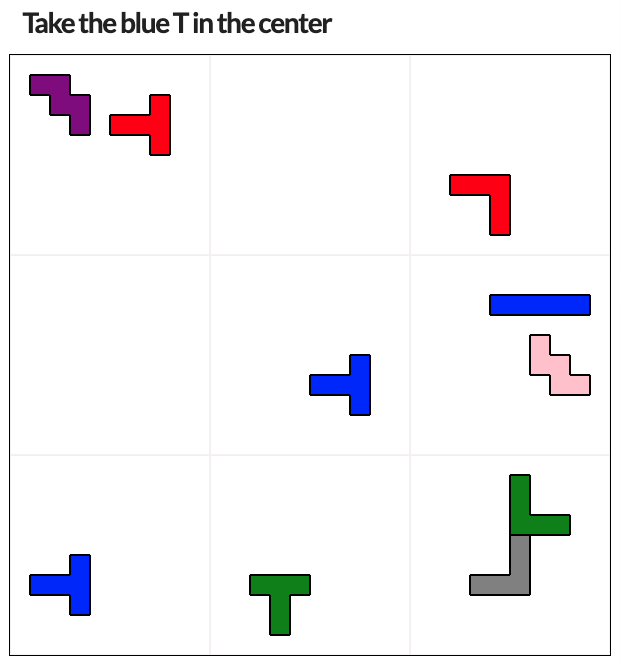}
    \end{center}
    \vspace{-0.3cm}
    \caption{A sample board with the target piece \texttt{(T,blue,center)} for this expression type.}
    \label{fig:ut_7}
    \vspace{-0.3cm}
\end{figure}

Mention the \textbf{color} excludes some, but not all. Mention the \textbf{shape} excludes some, but not all. Mention the \textbf{position} excludes the rest. We add one distractor that has the same color and shape (but a differen position) and one distractor that has the same color (but a different shape and position) and any other distractors. This requires at least 3 distractors.

\section{Model Details}

\subsection{LSTM}

Parameters: $53,201,327$ ($127$ MB) \\
GPU RAM: $3,423$ MiB (Batch $24$; \textsc{VE})

\begin{table}[h]
    \centering
    \small
    \begin{tabular}{|l|r|}
        \hline
         lstm\_hidden\_size & 1024 \\
         word\_embedding\_dim & 512 \\
         visual\_embedding\_dim & 512 \\
         dropout & 0.5 \\
         lr & $0.0003$ \\
         l2 & $0.0001$ \\
         gradient\_clip\_val & $10$ \\
         \hline
    \end{tabular}
    \caption{LSTM hyperparameters}
    \label{tab:lstm_hyperparams}
    \vspace{-0.7cm}
\end{table}

\subsection{Classifier}

Classes: $1,689$\\
Parameters: $30,234,718$ ($120$ MB)\\
GPU RAM: $11,063$ MiB (Batch $24$; \textsc{VSE})

\begin{table}[h]
    \centering
    \small
    \begin{tabular}{|l|r|}
        \hline
         d\_model & 512 \\
         visual\_embedding\_dim & 512 \\
         lr & $0.001$ \\
         l2 & $0.01$ \\
         layer\_norm & $0.00001$ \\
         gradient\_clip\_val & $10$ \\
         \hline
    \end{tabular}
    \caption{Linear model hyperparameters}
    \label{tab:linear_hyperparams}
    \vspace{-0.7cm}
\end{table}

\subsection{Transformer}

Parameters: $37,402,090$ ($149$ MB)\\
GPU RAM: $10,871$ MiB (Batch $24$; \textsc{VSE})

\begin{table}[h]
    \centering
    \small
    \begin{tabular}{| l | r| }
        \hline
         d\_model & 512 \\
         word\_embedding\_dim & 512 \\
         visual\_embedding\_dim & 512 \\
         nhead & 4 \\
         num\_encoder\_layers & 3 \\
         num\_decoder\_layers & 3 \\
         dim\_feedforward & 1024 \\
         dropout & 0.2 \\
         lr\_initial & $\text{d\_model}^{-0.5}$ \\
         l2 & $0.0001$ \\
         layer\_norm & $0.00001$ \\
         gradient\_clip\_val & $10$ \\
        \hline
    \end{tabular}
    \caption{Transformer hyperparameters}
    \label{tab:transformer_hyperparams}
    \vspace{-0.3cm}
\end{table}

\section{Error Analysis}
\label{appendix:qualitative_analysis}

Our best Transformer+\textsc{vse} model predicts $1,119$ of $12,436$ evaluation expressions wrong meaning that the prediction does not match the reference perfectly. Here 425 errors (213 data, 77 ho-pos, 65 ho-color, 70 ho-uts) are expression predictions where the target piece is the one for which the board was initially designed for and 694 (all data) are cases where we picked an additional target randomly.

\subsection{First-class errors}

\begin{table}[h]
    \centering
    \small
    \begin{tabular}{|l|r|r|r|r|}
        \hline
         Error types & color & shape & pos & ungram.\\
         \hline
         data & 4 & 5 & 180 & 24\\
         ho-color & 2 & 13 & 47 & 5 \\
         ho-pos & 5 & 5 & 58 & 9 \\
         ho-uts & 4 & 15 & 43 & 9 \\
         \hline
    \end{tabular}
    \caption{The error types for the first-class errors}
    \label{tab:error_first_class}
    \vspace{-0.1cm}
\end{table}

For the 425 first-class errors 213 of the errors are related to cases where the model mentions more properties of the target piece, although this would be unnecessary. In 47 cases the model produces an expressions that is not necessarily incorrect, but not grammatical.

\subsection{Second-class errors}

\begin{table}[h]
    \centering
    \small
    \begin{tabular}{|l|r|r|r|r|}
        \hline
         Error types & color & shape & pos & ungram.\\
         \hline
         data & 30 & 56 & 524 & 90\\
         \hline
    \end{tabular}
    \caption{The error types for the second-class errors}
    \label{tab:error_second_class}
    \vspace{-0.1cm}
\end{table}

For the 694 second-class errors 179 of the errors are related to cases where the model mentions in addition the position, color or shape of the target piece, although this would be unnecessary. In 90 cases the model produces an expressions that is not necessarily incorrect, but not grammatical.

\clearpage

\twocolumn[\subsection{First-class error examples (intended target pieces)}]

\includegraphics[width=.99\textwidth]{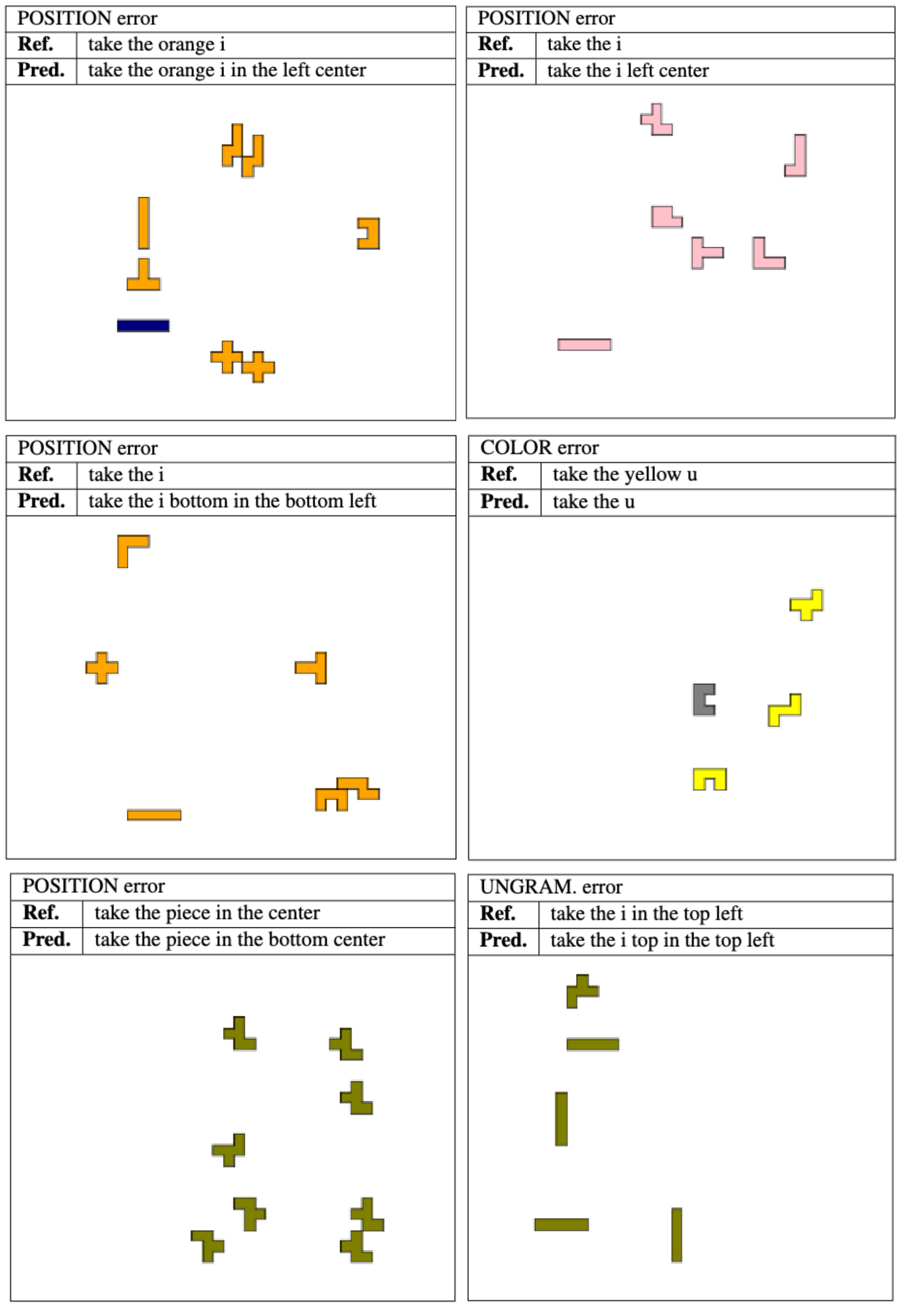}

\clearpage

\twocolumn[\subsection{Second-class error examples (extra target pieces)}]

\includegraphics[width=.99\textwidth]{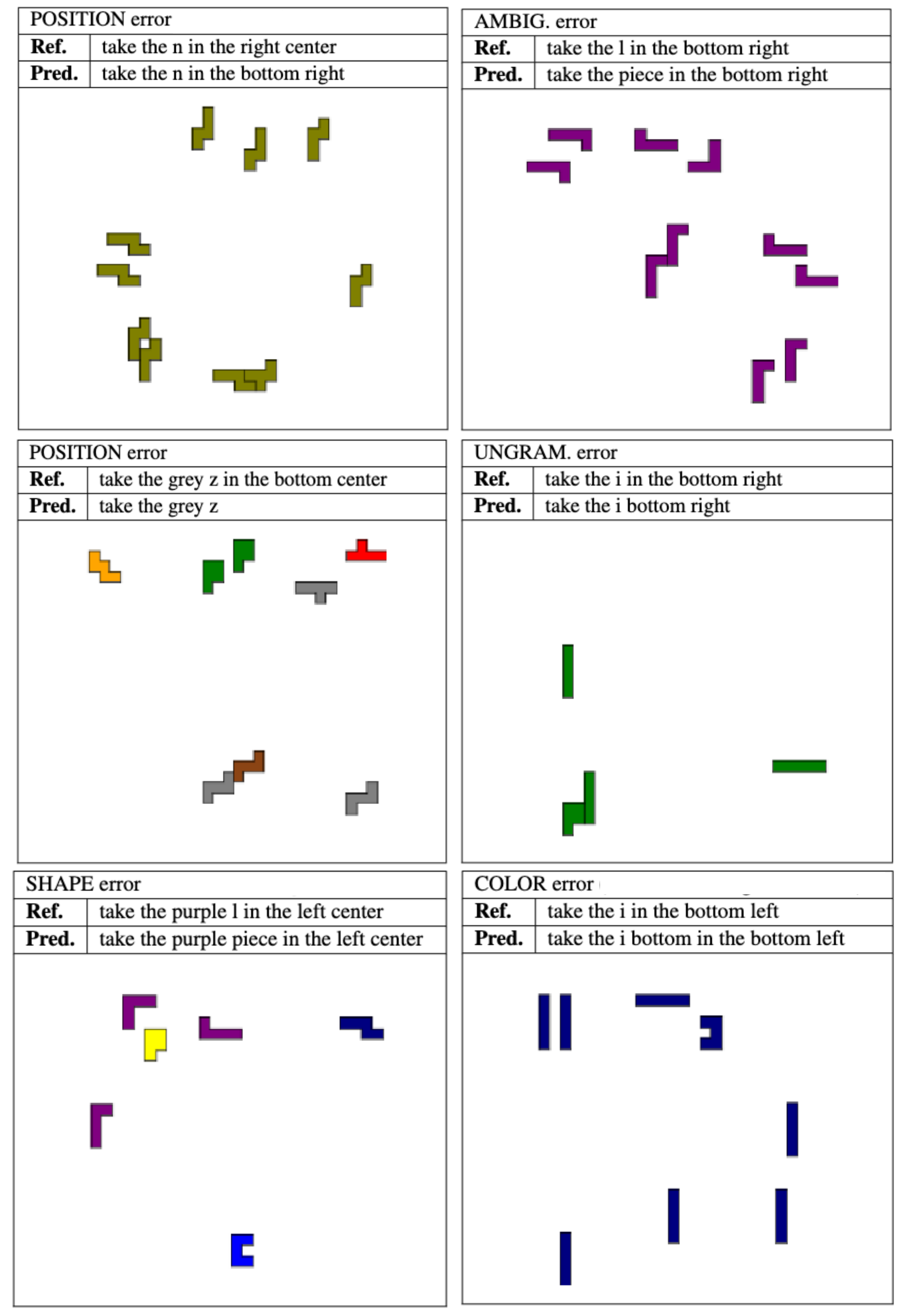}

\clearpage

\end{document}